%% file: ms.tex
\documentclass[runningheads]{llncs}
\usepackage[T1]{fontenc}
\usepackage{amssymb} 
\usepackage{amsmath} 
\usepackage{graphicx} 
\usepackage{caption}
\usepackage{subcaption}
\usepackage{comment}
\usepackage[misc]{ifsym} 
\usepackage{enumitem} 
\usepackage{float}
\usepackage{booktabs} 

\usepackage{mathtools}
\usepackage{bm} 
\usepackage{siunitx} 
\usepackage{dsfont} 
\usepackage{xspace} 
\input{z_basic-math}
\input{z_basic-ml}
\input{z_ml-interpretable}

\usepackage{tikz,forest}
\usetikzlibrary{arrows, fit, shapes, positioning, arrows.meta}
\tikzset{
    >=stealth',
    punkt/.style={
    rectangle,
    rounded corners,
    draw=black, very thick,
    text width=6.5em,
    minimum height=2em,
    text centered},
    pil/.style={
    ->,
    thick,
    shorten <=2pt,
    shorten >=2pt,},
    treenode/.style = {shape=rectangle, rounded corners,
                     draw, align=center},
     root/.style    = {treenode, font=\normalsize},
    env/.style      = {treenode, font=\normalsize},
    dummy/.style    = {circle,draw}
}

\usepackage[hidelinks]{hyperref}
\usepackage{color}

\usepackage{centernot}
\newcommand{\independent}{\perp\mkern-9.5mu\perp}
\newcommand{\notindependent}{\centernot{\independent}}

\spnewtheorem{result}{Result}[subsection]{\bfseries}{\itshape}
\spnewtheorem{negresult}[result]{Negative Result}{\itshape}{\itshape}

\makeatletter
\newcommand{\printfnsymbol}[1]{%
  \textsuperscript{\@fnsymbol{#1}}%
}
\makeatother

\begin{document}

%
\title{A Guide to Feature Importance Methods for Scientific Inference}
%
%
\author{Fiona Katharina Ewald\inst{1,2}\orcidID{0009-0002-6372-3401} \and 
Ludwig Bothmann\inst{1,2}\orcidID{0000-0002-1471-6582} \and
Marvin N. Wright \inst{4,5,6}\orcidID{0000-0002-8542-6291} \and
Bernd Bischl\inst{1,2}\orcidID{0000-0001-6002-6980} \and
Giuseppe Casalicchio\thanks{equal contribution as senior authors}\inst{,1,2}\orcidID{0000-0001-5324-5966}\,\Letter\, \and
Gunnar König\printfnsymbol{1}\inst{,3}\orcidID{0000-0001-6141-4942}}
%
\authorrunning{F. K. Ewald et al.}
%
\institute{Department of Statistics, LMU Munich, Munich, Germany\\
\Letter\,\email{Giuseppe.Casalicchio@stat.uni-muenchen.de} \and
Munich Center for Machine Learning (MCML), Munich, Germany \and
Department of Computer Science and Tübingen AI Center, University of Tübingen, Tübingen, Germany \and
Leibniz Institute for Prevention Research \& Epidemiology -- BIPS, Bremen, Germany \and
University of Bremen, Bremen, Germany \and
University of Copenhagen, Copenhagen, Denmark}

\maketitle              

\begin{abstract}
While machine learning (ML) models are increasingly used due to their high predictive power, their use in understanding the data-generating process (DGP) is limited.
Understanding the DGP requires insights into feature-target associations, which many ML models cannot directly provide due to their opaque internal mechanisms.
Feature importance (FI) methods provide useful insights into the DGP under certain conditions. 
Since the results of different FI methods have different interpretations, selecting the correct FI method for a concrete use case is crucial and still requires expert knowledge. 
This paper serves as a comprehensive guide to help understand the different interpretations of global FI methods.
Through an extensive review of FI methods and providing new proofs regarding their interpretation, we facilitate a thorough understanding of these methods and formulate concrete recommendations for scientific inference. 
We conclude by discussing options for FI uncertainty estimation and point to directions for future research aiming at full statistical inference from black-box ML models.

\keywords{Feature Importance  \and Model-agnostic Interpretability \and Interpretable ML}
\end{abstract}


\section{Introduction}\label{intro}

Machine learning (ML) models have gained widespread adoption, demonstrating their ability to model complex dependencies and make accurate predictions \cite{jordan2015machine}.
Besides accurate predictions, practitioners and scientists are often equally interested in understanding the data-generating process (DGP) to gain insights into the underlying relationships and mechanisms that drive the observed phenomena \cite{shmueli2010explain}. 
Since analytical information regarding the DGP is mostly unavailable, one way is to analyze a predictive model as a surrogate. 
Although this approach has potential pitfalls, it can serve as a viable alternative for gaining insights into the inherent patterns and relationships within the observed data, particularly when the generalization error of the ML model is small \cite{Molnar2022}.
Regrettably, the complex and often non-linear nature of certain ML models renders them opaque, presenting a significant challenge in understanding them.

A broad range of interpretable ML (IML) methods have been proposed in the last decades \cite{Covert2021,Guidotti2018}. These include \textit{local} techniques that only explain one specific prediction as well as \textit{global} techniques that aim to explain the whole ML model or the DGP; \textit{model-specific} techniques that require access to model internals (e.g., gradients) as well as \textit{model-agnostic} techniques that can be applied to any model; and \textit{feature effects} methods, which reflect the change in the prediction depending on the value of the feature of interest (FOI), as well as \textit{feature importance} (FI) methods, which assign an importance value to each feature depending on its influence on the prediction performance.
We argue that in many scenarios, analysts are interested in reliable statistical, population-level inference regarding the underlying DGP \cite{molnar2023relating,Williamson2023}, instead of ``simply'' explaining the model's internal mechanisms or heuristic computations whose exact meaning regarding the DGP is at the very least unclear or not explicitly stated at all.
If an IML technique is used for such a purpose, it should ideally be clear, what property of the DGP is computed, and, as we nearly always compute on stochastic and finite data, how variance and uncertainty are handled. 
The relevance of IML in the context of scientific inference has been recognized in general \cite{shmueli2010explain} as well as in specific subfields, e.g., in medicine \cite{caruana2015intelligible} or law \cite{doshi2017accountability}.
Krishna et al. \cite{Krishna2022} illustrate the disorientation of practitioners when choosing an IML method. 
In their study, practitioners from both industry and science were asked to choose between different IML methods and explain their choices. 
The participants predominantly based their choice on superficial criteria such as publication year or whether the method's outputs align with their prior intuition,
highlighting the absence of clear guidelines and selection criteria for IML techniques.

\begin{figure}[t]
\centering
\begin{subfigure}{.36\textwidth}
  \centering
  \includegraphics[clip, trim=0.6cm 0.7cm 0cm 0cm, width=\linewidth]{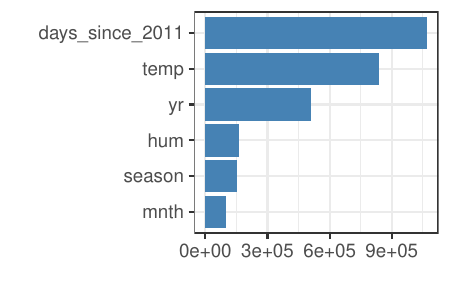}
  \caption{PFI}
  \label{fig:obese_rf_fi}
\end{subfigure}
\begin{subfigure}{.36\textwidth}
  \centering
  \includegraphics[clip, trim=0.6cm 0.7cm 0cm 0cm, width=\linewidth]{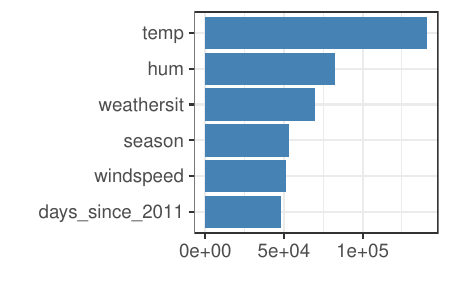}
  \caption{LOCO}
  \label{fig:obese_loco}
\end{subfigure}
\caption{Six most important features following (a)~PFI and (b)~LOCO.}
\label{fig:rf_intro}
\end{figure}

\paragraph{Motivating Example.}
The well-known ``bike sharing'' data set \cite{FanaeeT2013} includes 731 observations and 12 features corresponding to, e.g., weather, temperature, wind speed, season, and day of the week.
Suppose a data scientist is not only interested in achieving accurate predictions of the number of bike rentals per day but also in learning about the DGP to identify how the features are associated with the target. 
She trains a default random forest (RF, test-RMSE: 623, test-$R^2$: 0.90), and for analyzing the DGP, she decides to use two FI methods: permutation feature importance (PFI) and leave-one-covariate-out (LOCO) with L2 loss (details on these follow in Sections \ref{subsec:rules:univariate} and \ref{subsec:rules:refitting}). 
Unfortunately, she obtains somewhat contradictory results -- shown in Figure \ref{fig:rf_intro}. 
The methods produce results that agree on using temperature (\texttt{temp}), season (\texttt{season}), the number of days elapsed since the start of data collection in 2011 (\texttt{days\_since\_2011}), and humidity (\texttt{hum}) as part of the top 6 most important features, but the rankings of these features differ across different methods. 
She is unsure which feature in the DGP is the most important one, what the disagreement of the FI methods means, and, most importantly, what she can confidently infer from the results about the underlying DGP. 
We will address her questions in the following sections.

\paragraph{Contributions and Outline.} 
This paper assesses the usefulness of several FI methods for gaining insight into associations between features and the prediction target in the DGP.
Our work is the first concrete and encompassing guide for global, loss-based, model-agnostic FI methods directed toward researchers who aim to make informed decisions on the choice of FI methods for (in)dependence relations in the data.
The literature review in Section \ref{related_work} highlights the current state-of-the-art and identifies a notable absence of guidelines.
Section \ref{sec:scope_DGP_properties} determines the type of feature-target associations within the DGP that shall be analyzed with the FI methods.
In Section \ref{subsec:rules:univariate}, we discuss methods that remove features by perturbing them; in Section \ref{subsec:rules:multivariate} methods that remove features by marginalizing them out; and in Section \ref{subsec:rules:refitting} methods that remove features by refitting the model without the respective features.
In each of the three sections, we first briefly introduce the FI methods, followed by an interpretation guideline according to the association types introduced in Section \ref{sec:scope_DGP_properties}.
At the end of each section, our results are stated mathematically, with some proofs provided in Appendix \ref{appendix_additional_proofs}.
We return to our motivational example and additionally illustrate our theoretical results in a simulation study in Section \ref{sec_example} and formulate recommendations and practical advice in Section \ref{recommendations}.
We mainly analyze the \textit{estimands} of the considered FI, but it should be noted that the interpretation of the \textit{estimates} comes with additional challenges.
Hence, we briefly discuss approaches to measure and handle their uncertainty in Section \ref{uncertainty} and conclude in Section \ref{conclusion} with open challenges.



\section{General Notation}
Let $\D = \Dset$ be a data set of $n$ observations, which are sampled i.i.d. from a $p$-dimensional feature space $\Xspace = \Xspace_1 \times \ldots \times \Xspace_p$ and a target space $\Yspace$.
The set of all features is denoted by $P = \{1,\ldots,p\}$.
The realized feature vector is $\xi = (x^{(i)}_1, \ldots, x^{(i)}_p)^\top$, $i \in \{1, \ldots, n\}$,  where $\yv = \yvec$ are the realized labels. The associated random variables are $X=(X_1, \ldots, X_p)^\top$ and $Y$, respectively.
Marginal random variables for a subset of features $S \subseteq P$ are denoted by $X_S$.
The complement of $S$ is denoted by $-S = P \setminus S$.
Single features and their complements are denoted by 
$j$ and $-j$, respectively. 
Probability distributions are denoted by $F$, e.g., $F_{Y}(Y)$ is the marginal distribution of $Y$. 
If two random vectors, e.g., feature sets $X_J$ and $X_K$, are unconditionally independent, we write  $X_J \independent X_K$; if they are unconditionally dependent, which we also call unconditionally associated, we write $X_J \notindependent X_K$.

We assume an underlying true functional relationship $\ftrue: \Xspace \rightarrow \Yspace$ that implicitly defines the DGP by $Y = \ftrue(X) + \epsilon$. 
It is approximated by an ML model $\fh: \Xspace \rightarrow \R^g$, estimated 
on training data $\D$.
In the case of a regression model, $\Yspace = \R$, and $g = 1$. 
If $\fh$ represents a classification model, $g$ is greater or equal to $1$: for binary classification (e.g., $\Yspace = \{0,1\}$), $g$ is $1$; for multi-class classification, it represents the $g$ decision values or probabilities for each possible outcome class.
The ML model $\fh$ is determined by the so-called learner or inducer $\mathcal{I}: \D \times \lambda \mapsto \fh$ that uses hyperparameters $\lambda$ to map a data set $\D$ to a model in the hypothesis space $\fh \in \Hspace$.
Given a loss function, defined by $L: \Yspace \times \R^g \rightarrow \R_{0}^+$, the risk function of a model $\fh$ is defined as the expected loss $\risk(\fh) = \E[L(Y,\fh(X))]$.

\section{Related Work}\label{related_work}
Several papers aim to provide a general overview of existing IML methods \cite{Covert2021,Das2020,Guidotti2018,Han2022}, but they all have a very broad scope and do not discuss scientific inference.
Freiesleben et al. \cite{Freiesleben2023} propose a general procedure to design interpretations for scientific inference and provide a broad overview of suitable methods. 
In contrast, we provide concrete interpretation rules for FI methods.
Hooker et al. \cite{Hooker2021} analyze FI methods based on the reduction of performance accuracy when the FOI is unknown.
We examine FI techniques and provide recommendations depending on different types of feature-target associations. 

This paper builds on a range of work that assesses how FI methods can be interpreted: 
Strobl et al. \cite{Strobl2008} extended PFI \cite{breiman2001random} for random forests by using the conditional distribution instead of the marginal distribution when permuting the FOI, resulting in the conditional feature importance (CFI); Molnar et al. \cite{molnar2023model} modified CFI to a model-agnostic version where the dependence structure is estimated by trees;
König et al. \cite{Koenig2021} generalize PFI and CFI to a more general family of FI techniques called relative feature importance (RFI) and assess what insight into the dependence structure of the data they provide; 
Covert et al. \cite{Covert2020} derive theoretical links between Shapley additive global importance (SAGE) values and properties of the DGP; 
Watson and Wright \cite{Watson2021} propose a CFI based conditional independence test;
Lei et al. \cite{Lei2018} introduce LOCO and are among the first to base FI on hypothesis testing; 
Williamson et al. \cite{Williamson2023} present a framework for loss-based FI methods based on model refits, including hypothesis testing;
and Au et al. \cite{Au2022} focus on FI methods for groups of features instead of individual features, such as leave-one-group-out importance (LOGO). 

In addition to the interpretation methods discussed in this paper, other FI approaches exist. 
Another branch of IML deals with variance-based FI methods aimed at the FI of an ML model and not necessarily regarding the DGP, as they only use the prediction function of an ML model without considering the ground truth.
For example, the feature importance ranking measure (FIRM) \cite{Zien2009} uses a feature effect function and defines the standard deviation as an importance method. 
A similar method by \cite{Greenwell2018} uses the standard deviation of the partial dependence (PD) function \cite{Friedman2001} as an FI measure. 
The Sobol index \cite{Sobol1993sensitivity} is a more general variance-based method based on a decomposition of the prediction function into main effects and high-order effects (i.e., interactions) and estimates the variance of each component to quantify their importance \cite{owen2013variance}. 
Lundberg et al. \cite{lundberg5consistent} introduced the SHAP summary plot as a global FI measure based on aggregating local SHAP values \cite{lundberg2017unified}, which are defined only regarding the prediction function without considering the ground truth.
\section{Feature-Target Associations}\label{sec:scope_DGP_properties}
When analyzing the FI methods, we focus on whether they provide insight into (conditional) (in)dependencies between a feature $X_j$ and the prediction target $Y$. More specifically, we are interested in understanding whether they provide insight into the following relations:
\begin{enumerate}[label = (\textit{A\arabic*}), leftmargin=*]
\item Unconditional association ($X_j \notindependent Y$).\label{question:unconditional} 
\item Conditional association ... \label{question:conditional-general}
\begin{enumerate}[label=(\textit{A\arabic{enumi}\alph*}), leftmargin=*]
    \item ... given all remaining features $X_{-j}$  ($X_j \notindependent Y \, | \, X_{-j}$). \label{question:conditional}
    \item ... given any user-specified set $X_G,$ G $\subset P\backslash \{j\}$  ($X_j \notindependent Y \, | \, X_{G}$).  \label{question:relative}
\end{enumerate}
\end{enumerate}
An unconditional association \ref{question:unconditional} indicates that a feature $X_j$ provides information about $Y$, i.e., knowing the feature on its own allows us to predict $Y$ better; if $X_j$ and $Y$ are independent, 
this is not the case.
On the other hand, a conditional association \ref{question:conditional-general} with respect to (w.r.t.) a set $S \subseteq P\backslash \{j\}$ indicates that $X_j$ provides information about $Y$, even if \textit{we already know} $X_S$. 
When analyzing the suitability of the FI methods to gain insight into \ref{question:unconditional}-\ref{question:relative}, it is important to consider that
no FI score can simultaneously provide insight into more than one type of association. 
In supervised ML, we are often interested in the conditional association between $X_j$ and $Y$, given $X_{-j}$ \ref{question:conditional}, i.e., predicting $Y$ better if we are given information regarding all other features.

For example, given measurements of several biomarkers and a disease outcome, a doctor may not only be interested in a well-performing black-box prediction model based on all biomarkers but also in understanding which biomarkers are associated with the disease \ref{question:unconditional}. 
Furthermore, the doctor may want to understand whether measuring a biomarker is strictly necessary for achieving optimal predictive performance \ref{question:conditional} and to understand whether a set of other biomarkers $G$ can replace the respective biomarker \ref{question:relative}.

Example \ref{example:xor-independence} shows that conditional association does not imply unconditional association (\ref{question:conditional-general} $\not \Rightarrow$ \ref{question:unconditional}). Additionally, unconditional association does not imply conditional association, as Example \ref{example:chain-independence} demonstrates (\ref{question:unconditional} $\not \Rightarrow$ \ref{question:conditional-general}).
\begin{example}\label{example:xor-independence}
    Let $X_1, X_2 \sim Bern(0.5)$ be independent features and $Y := X_1 \oplus X_2$ (where $\oplus$ is the XOR operation). 
    Then, all three features are pairwise independent, but $X_1$ and $X_2$ together allow us to predict $Y$ perfectly.
\end{example}
\begin{example}\label{example:chain-independence}
Let $Y := X_1$ with $X_1 \sim N(0,1)$ and $X_2 := X_1 + \epsilon_2$ with $\epsilon_2 \sim N(0, 0.1)$.
Although $X_2$ provides information about $Y$, all of this information is also contained in $X_1$. Thus, $X_2$ is unconditionally associated with $Y$ but conditionally independent from $Y$ given $X_1$. 
\end{example}
Furthermore, conditional (in)dependence w.r.t. one feature set does not imply (in)dependence w.r.t. another, e.g., \ref{question:conditional} $\not \Leftrightarrow$ \ref{question:relative}. This is demonstrated by adding unrelated features to the DGP and the conditioning set, as shown in Examples \ref{example:xor-independence} and \ref{example:chain-independence}. 

\section{Methods Based on Univariate Perturbations}
\label{subsec:rules:univariate}

Methods based on univariate perturbations quantify the importance of a feature of interest (FOI) by comparing the model's performance before and after replacing the FOI $X_j$ with a perturbed version $\Tilde{X}_j$ (permuted observations):
\begin{align}
    \text{FI}_j = \E\left[L\left(Y, \fh(\Tilde{X}_{j}, X_{-j})\right)\right] - \E\left[L\left(Y, \fh(X)\right)\right]. \label{PFI_formula}
\end{align}
The idea behind this approach is that if perturbing the feature increases the prediction error, the feature should be important for $Y$. 
Below, we discuss the three methods PFI (Section \ref{sec_PFI_interpretation}), CFI (Section \ref{sec_CFI_interpretation}), and RFI (Section \ref{subsec:rfi}) differing in their perturbation scheme:
Perturbation in PFI \cite{breiman2001random,Fisher2019} preserves the feature's \textit{marginal distribution} while destroying all dependencies with other features $X_{-j}$ and the target $Y$, i.e.,
\begin{align}
    \Tilde{X}_{j}\sim F_{X_j}(X_j) \text{ and } \Tilde{X}_j \independent (X_{-j}, Y);
\end{align}
CFI \cite{Strobl2008} perturbs the FOI while preserving its dependencies with \textit{the remaining features}, i.e.,
\begin{align}
    \Tilde{X}_j \sim F_{X_j\,|\,X_{-j}}(X_j\,|\,X_{-j}) \text{ and } \Tilde{X}_j \independent Y\, |\, X_{-j};
\end{align}
RFI \cite{Koenig2021} is a generalization of PFI and CFI since the perturbations preserve the dependencies with \textit{any user-specified set} $G$, i.e.,
\begin{align}
    \Tilde{X}_{j}\sim F_{X_j\,|\, X_G}(X_j\,|\, X_G) \text{ and } \Tilde{X}_j \independent Y,X_{P \backslash (G \cup \{j\})}\, |\, X_G.
\end{align}
To indicate on which set $G$ the perturbation of $j$ is conditioned, we denote RFI$_j^G$. 
We obtain PFI by setting $G=\emptyset$ and CFI by setting $G=-j$.
As will be shown, the type of perturbation strongly affects which features are considered relevant.
\subsection{Permutation Feature Importance (PFI)} \label{sec_PFI_interpretation}
\subsubsection{Insight into $X_j \notindependent Y$ \ref{question:unconditional}:}
Non-zero PFI does not imply an unconditional association with $Y$ (Negative Result \ref{negresult:pfi-extrapolation}).
In the proof of Negative Result \ref{negresult:pfi-extrapolation}, we construct an example where the PFI is non-zero because the perturbation breaks the dependence between the features (and not because of an unconditional association with $Y$). 
Based on this, one may conjecture that unconditional feature independence is a sufficient assumption for non-zero PFI to imply an unconditional association with $Y$; however, this is not the case, as Negative Result \ref{negresult:pfi-interactions} demonstrates.
For non-zero PFI to imply an unconditional association with $Y$, the features must be independent \textit{conditional on $Y$} instead (Result \ref{result:non-zero-pfi}).

Zero PFI does not imply independence between the FOI and the target (Negative Result \ref{negresult:zero-pfi-l2-fallacy}). 
Suppose the model did not detect the association, e.g., because it is a suboptimal fit or because the loss does not incentivize the model to learn the dependence. PFI may be zero in that case, although the FOI is associated with $Y$. 
In the proof of Negative Result \ref{negresult:zero-pfi-l2-fallacy}, we demonstrate the problem for L2 loss, where the optimal prediction is the conditional expectation (and thus neglects dependencies in higher moments). 
For cross-entropy optimal predictors and given feature independence (both with and without conditioning on $Y$), zero PFI implies unconditional independence with $Y$ (Result \ref{result:zero-pfi-ce}).

\subsubsection{Insight into $X_j \notindependent Y$ conditional on $X_G$ or $X_{-j}$ \ref{question:conditional-general}:}
PFI relates to unconditional (in)dependence and, thus, is not suitable for insight into conditional (in)dependence (see Section \ref{sec:scope_DGP_properties}).
%
%
%
\begin{result}[PFI Interpretation] For non-zero PFI, it holds that
\label{result:non-zero-pfi}
\begin{align}
    (X_{j} \independent X_{-j}\, | \,Y) \land (\text{PFI}_j \neq 0) \quad \Rightarrow \quad X_j \notindependent Y.
\end{align}
For cross-entropy loss and the respective optimal model, \label{result:zero-pfi-ce}
\begin{align}
    (X_j \independent X_{-j}) \land (X_j \independent X_{-j}\, |\, Y) \land (\text{PFI}_j = 0) \quad \Rightarrow \quad X_j \independent Y.
\end{align}
\end{result}
\begin{proof} 
  The first implication directly follows from Theorem 1 in \cite{Koenig2021}.
  The second follows from the more general Result \ref{result:zero-rfi-ce}.\qed
\end{proof}
\begin{negresult}
\label{negresult:pfi-extrapolation}
$\text{PFI}_j \neq 0 \: \not \Rightarrow \: X_j \notindependent Y.$
\end{negresult}
\begin{proof}[Counterexample]
Let $Y, X_1 \sim N(0,1)$ be two independent random variables, $X_2:= X_1$, and 
the prediction model $\fh(x) = x_1 - x_2$. 
It is simple to calculate that this model has expected L2 loss of 1, as
$\E[L(Y, X_1-X_2)] = \E[Y^2] = 1$.
Now let $\Tilde{X}_1$ be the perturbed version of $X_1$ ($\Tilde{X}_1 \sim F_{X_j}(X_1)$), and $\Tilde{X}_1 \independent (Y, X_2)$. 
The expected L2 loss under perturbation now is
$\E[(Y - (\Tilde{X}_1-X_2))^2] = \text{Var}(Y - \Tilde{X}_1 + X_2) = 3$, which implies PFI$_1 = 2$. So PFI$_1$ is non-zero, but $X_1 \independent Y$.
\qed
\end{proof}
\begin{negresult}
\label{negresult:pfi-interactions}
    $(X_j \independent X_{-j}) \land (\text{PFI}_j \neq 0) \: \not \Rightarrow \: X_j \notindependent Y.$
\end{negresult}
\begin{proof}[Counterexample]
Let $X_1, X_2 \sim Bern(0.5)$ with $X_1 \independent X_2$, and $Y:= X_1 \oplus X_2$, where $\oplus$ is XOR. Consider a perfect prediction model $\fh(X) = x_1 \oplus x_2$, and $\fh$ encodes the posterior probability for $Y=1$ (here, $Y$ can be only 0 or 1).
This model has a cross-entropy loss of 0, since $Y = \fh(X)$. 
Furthermore, it holds that $X_1 \independent Y$.
Again, let $\Tilde{X}_1$ be the perturbed version of $X_1$.
One can easily verify that $Y = (X_1 \oplus X_2) \independent (\tilde{X}_1 \oplus X_2) = \tilde{\hat{Y}}$ and $Y, \Tilde{\hat{Y}} \sim Bern(0.5)$. Thus, the prediction $\tilde{\hat{Y}}$ using the perturbed feature $\Tilde{X}_1$ assigns probability $1$ to the correct and wrong class with probability $0.5$ each. Thus, the cross-entropy loss for the perturbed prediction is non-zero (actually, positive infinity), and PFI$_j \neq 0$. 
\qed
\end{proof}
\begin{negresult}
\label{negresult:zero-pfi-l2-fallacy}
FI$_j=0 \: \not \Rightarrow \: X_j \independent Y\, |\, X_G$ for any $G \subseteq P \backslash \{j\}$, even if the model is L2-optimal.\\
NB: This result holds not only for PFI but also for any FI method based on univariate perturbations, including PFI, CFI, and RFI (Equation \ref{PFI_formula}).%
\end{negresult}
\begin{proof}[Counterexample]
If a model does not rely on a feature $X_j$, FI$_j = 0$.
We construct an example where $\fh$ is L2-optimal but does not rely on the feature $X_1$, which is dependent with $Y$ conditional on any set $G \subseteq P \backslash \{j\}$.
Let $Y\,|\,X_1, X_2 \sim N(X_2, X_1)$ with $X_1, X_2 \sim N(0,1)$ and $X_1 \independent X_2$.
Then, $Y$ is dependent with $X_1$ conditional on any set $G \subseteq P \backslash \{1\}$: Here, $G$ could either be $G = \emptyset$ or $G = \{2\}$. Now, for small $X_1$, extreme values of $Y$ are less likely than for $X_1=100$, irrespective of whether we know $X_2$.
Now consider $\fh(x) = x_2$. $\fh$ is L2-optimal since $\E[Y|X]=X_2$, but $\fh$ does not depend on $X_1$.
 \qed
\end{proof}

\subsection{Conditional Feature Importance (CFI)} \label{sec_CFI_interpretation}
\subsubsection{Insight into $X_j \notindependent Y\, |\, X_{-j}$ \ref{question:conditional}:}
Since CFI preserves associations between features, non-zero CFI implies a conditional dependence on $Y$, even if the features are dependent (Result \ref{propCFI}).
The converse generally does not hold, so Negative Result \ref{negresult:zero-pfi-l2-fallacy} also applies to CFI. 
However, for cross-entropy optimal models, zero CFI implies conditional independence (Result \ref{result:cfi-independence-ce}).
\subsubsection{Insight into $X_j \notindependent Y$ \ref{question:unconditional} and  $X_j \notindependent Y\, |\, X_G$ \ref{question:relative}:}
Since CFI provides insight into conditional dependence \ref{question:conditional}, it follows from Section \ref{sec:scope_DGP_properties} that CFI is not suitable to gain insight into \ref{question:unconditional} and \ref{question:relative}. 
\begin{result}[CFI interpretation]
\label{propCFI} For CFI, it holds that
\begin{align}
    \text{CFI}_j \neq 0 \quad \Rightarrow \quad X_j \notindependent Y \, |\, X_{-j}
\end{align}
For cross-entropy optimal models, the converse holds as well.\label{result:cfi-independence-ce}
\end{result}
\begin{proof}
    The first equation follows from Theorem 1 in \cite{Koenig2021}. 
    The second follows from the more general Result \ref{result:zero-rfi-ce}. \qed
\end{proof}
\subsection{Relative Feature Importance (RFI)} \label{subsec:rfi}

\subsubsection{Insight into $X_j \notindependent Y\, |\, X_G$ \ref{question:relative}:}
Result \ref{propRFi} generalizes Results \ref{result:non-zero-pfi} and \ref{propCFI}. 
While PFI and CFI are sensitive to dependencies conditional on \textit{no} or \textit{all} remaining features, RFI is sensitive to conditional dependencies w.r.t. \textit{a user-specified feature set $G$}.
Nevertheless, we must be careful with our interpretation if features are dependent. 
RFI may be non-zero even if the FOI is not associated with the target (Negative Result \ref{negresult:rfi-extrapolation}).
In general, zero RFI does not imply independence (Negative Result \ref{negresult:zero-pfi-l2-fallacy}). 
Still, for cross-entropy optimal models and under independence assumptions, insight into conditional independence w.r.t. $G$ can be gained (Result \ref{result:zero-rfi-ce}).
\subsubsection{Insight into $X_j \notindependent Y$ \ref{question:unconditional} and $X_j \notindependent Y\, |\, X_{-j}$\ref{question:conditional}:}
If features are conditionally independent given $Y$, setting $G$ to $\emptyset$ (yielding PFI) enables insight into unconditional dependence. Setting $G$ to $-j$ (yielding CFI) enables insight into the conditional association given all other features. 
\begin{result}[RFI interpretation]\label{propRFi}
For $R=P\backslash (G \cup \{ j\})$, it holds that
\begin{align}
    (X_j \independent X_R \, |\, X_G, Y) \land (\text{RFI}_j^G \neq 0) \quad \Rightarrow \quad X_j \notindependent Y \, |\,\, X_G.
\end{align}
For cross-entropy optimal predictors 
and $G \subseteq P \backslash \{j\}$, it holds that \label{result:zero-rfi-ce}
\begin{align}
    (X_j \independent X_R \, |\, X_G, Y) \land (X_j \independent X_R \, |\, X_G) \land (\text{RFI}_j^G = 0) \: \Rightarrow \: X_j \independent Y \, |\, X_G.
\end{align}
\end{result}
\begin{proof}
The first implication follows directly from Theorem 1 in \cite{Koenig2021}. 
The proof of the second implication 
can be found in Appendix \ref{proof:result:zero-rfi-ce}.
\end{proof}
\begin{negresult}
\label{negresult:rfi-extrapolation}
$RFI_j^G \neq 0 \: \not \Rightarrow \:  X_j \notindependent Y \, |\, X_G$.
\end{negresult}
\begin{proof}[Counterexample] 
Let $G = \emptyset$. Then, $RFI_j^G = PFI_j$ and $X_j \notindependent Y\, |\, X_G \Leftrightarrow X_j \notindependent Y$. Thus, the result directly follows from \ref{negresult:pfi-extrapolation}.
\qed
\end{proof}
\section{Methods Based on Marginalization}
\label{subsec:rules:multivariate}
In this section, we assess SAGE value functions (SAGEvf) and SAGE values \cite{Covert2020}. The methods remove features by marginalizing them out of the prediction function. The marginalization \cite{lundberg2017unified} is performed using either the conditional or marginal expectation. 
These so-called reduced models are defined as
\begin{align}
\begin{split}
    &\fh^m_S(x_S) = \E_{X_{-S}}\left[\fh(x_S, X_{-S})\right], \quad \text{ and}\\
    &\fh^c_S(x_S) = \E_{X_{-S}\,|\,X_S}\left[\fh(x_S, X_{-S})\, |\, X_S\right],
\end{split}
    \label{def_reduced_f}
\end{align} 
where $\fh^m$ is the marginal and $\fh^c$ is the conditional-sampling-based version and $\fh^m_{\emptyset}=\fh^c_{\emptyset}$ the average model prediction, e.g., $\E[Y]$ for an L2 loss optimal model and $\P(Y)$ for a cross-entropy loss optimal model. 
Based on these, SAGEvf quantify the change in performance that the model restricted to the FOIs achieves over the average prediction: 
\begin{align} 
v^{m/c}(S) = \E\left[L\left(Y, \fh^{m/c}_{\emptyset}\right)\right] - \E\left[L\left(Y, \fh^{m/c}_S(X_{S})\right)\right] 
\label{def_SAGEvf_S}
\end{align}
We abbreviate SAGEvf depending on the distribution used for the restricted prediction function (i.e., $\fh^m$ or $\fh^c$) with mSAGEvf ($v^m$) and cSAGEvf ($v^c$). 

SAGE values \cite{Covert2020} regard FI quantification as a cooperative game, where the features are the players, and the overall performance is the payoff. 
The surplus performance (surplus payoff) enabled by adding a feature to the model depends on which other features the model can already access (coalition).
To account for the collaborative nature of FI, SAGE values use Shapley values \cite{Shapley1951} to divide the payoff for the collaborative effort (the model’s performance) among the players (features).
SAGE values are calculated as the weighted average of the surplus evaluations over all possible coalitions $\SsubPnoj$:
\begin{align}
\phi^{m / c}_j(v) &= \frac{1}{p} \sum_{\SsubPnoj} \binom{p-1}{|S|}^{-1} \left( v^{m / c}(\Scupj) - v^{m / c}(S) \right), 
\end{align}
where the superscript in $\phi_j$ denotes whether the marginal $v^m(S)$ or conditional $v^c(S)$ value function is used. 

\subsection{Marginal SAGE Value Functions (mSAGEvf)}
\label{subsec:mSAGEvf}

\subsubsection{Insight into $X_j \notindependent Y$ \ref{question:unconditional}:}
Like PFI, mSAGE value functions use marginal sampling and break feature dependencies. mSAGEvf may be non-zero ($v^m(\{j\}) \neq 0$), although the respective feature is not associated with $Y$ (Negative Result \ref{negresult:mSAGEvf-extrapolation}).
While an assumption about feature independence was sufficient for PFI for insight into pairwise independence, this is generally not the case for mSAGEvf. 
The feature marginalization step may lead to non-zero importance for non-optimal models (Negative Result \ref{negresult:SAGEvf-suboptimal}).
Given feature independence and L2 or cross-entropy optimal models, a non-zero mSAGEvf implies unconditional association; the converse only holds for CE optimal models (Result \ref{mSAGEvf_int}).
\subsubsection{Insight into $X_j \notindependent Y$ conditional on $X_G$ or $X_{-j}$ \ref{question:conditional-general}:}
The method mSAGEvf does not provide insight into the dependence between the FOI and $Y$ (Negative Result \ref{negresult:mSAGEvf-extrapolation}) unless the features are independent and the model is optimal w.r.t. L2 or cross-entropy loss (Result \ref{mSAGEvf_int}). 
Then, mSAGEvf can be linked to \ref{question:unconditional} and, thus, is not suitable for \ref{question:conditional-general} (Section \ref{sec:scope_DGP_properties}).
\begin{result}[mSAGEvf interpretation]
\label{mSAGEvf_int}
For L2 loss or cross-entropy loss-optimal models (and the respective loss) and $(X_j \independent X_{-j})$, it holds that
\begin{align}
    v^m(\{j\}) \neq 0  \quad \Rightarrow \quad X_j \notindependent Y
\end{align}
For cross-entropy optimal predictors, the converse holds as well.
\end{result}
\begin{proof}
   The proof can be found in Appendix \ref{proof_eq_mSAGEvf}.
\end{proof}
\begin{negresult}\label{negresult:mSAGEvf-extrapolation}
$v^m(\{j\}) \neq 0 \: \not \Rightarrow \: \exists\, G \subseteq P \backslash \{j\} : X_j \notindependent Y\, |\, X_G$
\end{negresult}
\begin{proof}[Counterexample]
Let us assume the same DGP and model as in the proof of Negative Result \ref{negresult:pfi-extrapolation}.
In the setting, both the full model $\fh^m_1(x) = x_1 - x_2 = 0$ and $\fh^m_\emptyset = 0$ are optimal, but $\fh^m_1(x_1) = x_1 \neq 0$ is sub-optimal. 
Thus, $v^m(\{1\}) \neq 0$ (although $X_1 \independent Y$ for any $G \subseteq P \backslash \{j\}$). \qed
\end{proof}
\begin{negresult}\label{negresult:SAGEvf-suboptimal}
$(v(\{j\}) \neq 0) \land (X_j \independent X_{-j}) \: \not \Rightarrow \: X_j \notindependent Y$
\end{negresult}
\begin{proof}[Counterexample]
Let $X_1, Y \sim N(0,1)$, and let $X_{-1}$ be some (potentially multivariate) random variable, with $X_{-1} \independent Y$ and $X_{-1} \independent X_1$. 
Let $\fh(x) = x_1$ be the prediction model.
Then, $\fh^m_\emptyset = \fh^c_\emptyset = \E_X[X_1] = 0$ and $\fh^m_1(x_1) = \fh^c_1(x_1) = x_1$. 
Since the optimal prediction $\hat{Y}^* = \E[Y|X] = \E[Y] = 0$, the average prediction $\fh^m_\emptyset = \fh^c_\emptyset = 0$ is loss-optimal and $\fh^m_1(x_1) = \fh^c_1(x_1) = x_1$ is not loss-optimal.
Consequently, $v(\{1\}) \neq 0$ (although $X_1$ is independent of target and features). 
Notably, the example works both for $v^m$ and $v^c$. \qed
\end{proof}

\subsection{Conditional SAGE Value Functions (cSAGEvf)}
\label{subsec:cSAGEvf}

\subsubsection{Insight into $X_j \notindependent Y$ \ref{question:unconditional}:}
Like for mSAGEvf, model optimality w.r.t. L2 or cross-entropy loss is needed to gain insight into the dependencies in the data (Negative result \ref{negresult:SAGEvf-suboptimal}). However,  since cSAGEvf preserves associations between features, the assumption of independent features is not required to gain insight into unconditional dependencies (Result \ref{cSAGEvf_int}). 

\subsubsection{Insight into $X_j \notindependent Y$ conditional on $X_G$ or $X_{-j}$ \ref{question:conditional-general}:} 
Since cSAGEvf provide insight into \ref{question:unconditional}, they are unsuitable for gaining insight into \ref{question:conditional-general} (see Section \ref{sec:scope_DGP_properties}).
However, the difference between cSAGEvf for different sets, called surplus cSAGEvf (scSAGEvf$^G_j:=v^c(G \cup \{j\}) - v^c(G)$, where $G\subseteq P\backslash \{j\}$ is user-specified), provides insights into conditional associations (Result \ref{cSAGEvf_int}).  
\begin{result}[cSAGEvf interpretation]
\label{cSAGEvf_int}
For L2 loss or cross-entropy loss optimal models, it holds that: 
\begin{align}
   v^c(\{j\}) \neq 0\quad &\Rightarrow \quad X_j \notindependent Y\\
   \text{scSAGEvf}^G_j \neq 0 \quad &\Rightarrow \quad  X_j \notindependent Y\, |\, X_G
\end{align}
For cross-entropy loss, the respective converse holds as well.
\end{result}
\begin{proof}
The first implication (and the respective converse) follows from the second (and the respective converse) by setting $G=\emptyset$.
The second implication was proven in Theorem 1 in \cite{Luther2023}. For the converse, \cite{Covert2020} show that for cross-entropy optimal models $v^c(G \cup \{j\})-v^c(G) = I(Y,X_j|X_G)$; it holds that $I(Y,X_j|X_G) = 0 \Leftrightarrow X_j \independent Y\, |\, X_G$. \qed
\end{proof} 

\subsection{SAGE Values}
\label{subsec:sage}

Since non-zero cSAGEvf imply (conditional) dependence and cSAGE values are based on scSAGEvf of different coalitions, cSAGE values are only non-zero if a conditional dependence w.r.t. some conditioning set is present (see Result \ref{cSAGE_interpretation}).
\begin{result}
\label{cSAGE_interpretation}
Assuming an L2 or cross-entropy optimal model, the following interpretation rule for cSAGE values holds for a feature $X_j$: 
\begin{align}
    \phi^c_j(v) \neq 0 \quad &\Rightarrow \quad \exists\, \SsubP\backslash\{j\}: X_j \notindependent Y \, |\, X_S. 
\end{align}
For cross-entropy optimal models, the converse holds as well.
\end{result}
\begin{proof}
    The Proof can be found in Appendix \ref{proof_eq_cSAGE}.
\end{proof}
%
We cannot give clear guidance on the implications of mSAGE for \ref{question:unconditional}-\ref{question:relative} and leave a detailed investigation for future work.

\section{Methods Based on Model Refitting}
\label{subsec:rules:refitting}

This section addresses FI methods that quantify importance by removing features from the data and refitting the ML model. 
For LOCO \cite{Lei2018}, the difference in risk of the original model and a refitted model $\fhnj^r$ relying on every feature but the FOI $X_j$ is computed:
\begin{align}
    \text{LOCO}_j = \E\left[L\left(Y, \fhnj^r(X_{-j})\right)\right] - \E\left[L\left(Y, \fh(X)\right)\right],
\end{align}
where $\fhnj^r$ keeps the learner $\mathcal{I(\D, \lambda})$ fixed.\footnote{
In Eq.~\eqref{def_reduced_f}, we tagged the reduced models $\fh^m$ and $\fh^c$, indicating the type of marginalization. 
For refitting-based methods, we use the superscript $r$. 
}

Williamson et al. \cite{Williamson2023} generalize LOCO, as they are interested in not only one FOI but also in a feature set $S \subseteq P$.
As they do not assign an acronym, we from here on call it Williamson's Variable Importance Measure (WVIM):
\begin{align}
    \text{WVIM}_S = \E\left[L\left(Y, \fh_{-S}^r(X_{-S})\right)\right] - \E\left[L\left(Y, \fh(X)\right)\right].
\end{align}
Obviously, WVIM, also known as LOGO \cite{Au2022}, equals LOCO for $S=j$.
For $S=P$, the optimal refit reduces to the optimal constant prediction, e.g., for an L2-optimal model $\fh_{-S}^r(X_{-S})=\fh_{\emptyset}^r(x_{\emptyset}) = \E[Y]$ and for a cross-entropy optimal model $\fh_{\emptyset}^r(x_{\emptyset}) = \P(Y).$

\subsection{Leave-One-Covariate-Out (LOCO)} \label{sec_loco_interpretation}
For L2 and cross-entropy optimal models, LOCO is similar to $v^c(-j \cup j) - v^c(-j)$, with the difference that we do not obtain the reduced model by marginalizing out one of the features, but rather by refitting the model. 
As such, the interpretation is similar to the one of cSAGEvf (Result \ref{result:loco-interpretation}).
\begin{result}\label{result:loco-interpretation} 
    For an L2 or cross-entropy optimal model and the respective optimal reduced model $\fh^r_{-j}$, it holds that $LOCO_j \neq 0\: \Rightarrow \: X_j \notindependent Y\, |\, X_{-j}$. For cross-entropy loss, the converse holds as well.
\end{result}
\begin{proof}
    For cross-entropy and L2-optimal fits, the reduced model that we obtain from conditional marginalization behaves the same as the optimal refit (for cross-entropy loss $\fh^r_S = \fh^c_S = \P(Y|X_S)$, for L2 loss $\fh^r_S = \fh^c_S = \E[Y|X_S]$) \cite[Appendix B]{Covert2020} and thus LOCO$_j = v^c(j \cup -j) - v^c(-j)$. As such, the result follows directly from Result \ref{cSAGEvf_int}. 
\end{proof}

\subsection{WVIM as relative FI and Leave-One-Covariate-In (LOCI)}
For $S=j$, the interpretation is the same as for LOCO.
Another approach to analyzing the relative importance of the FOI is investigating the surplus WVIM (sWVIM$^{-G}_j$) for a group $G\subseteq P\backslash\{j\}$:
\begin{align}
    \text{sWVIM}^{-G}_j 
    &= \E\left[L\left(Y, \fh_{G}^r(X_{G})\right)\right] - \E\left[L\left(Y, \fh_{G \cup \{j\}}^r(X_{G \cup \{j\}})\right)\right].
\end{align}
It holds that sWVIM$^{-G}_j$ equals scSAGEvf$^G_j$, only differing in the way features are removed, so the interpretation is similar to the one of scSAGEvf. 
A special case results for $G = \emptyset$, i.e., the difference in risk between the optimal constant prediction and a model relying on the FOI only.
We refer to this (leaving-one-covariate-in) as LOCI$_j$.
For cross-entropy or L2-optimal models, the interpretation is the same as for cSAGEvf, since LOCI$_j = v^c(\{j\})$ (Result \ref{result:WVIM-interpretation}).
\begin{result}\label{result:WVIM-interpretation}
For L2 or cross-entropy optimal learners, it holds that 
\begin{align}
    \text{LOCI}_{j} \neq 0 \quad &\Rightarrow \quad X_j \notindependent Y, \qquad \text{ and}\\
    \text{sWVIM}^{-G}_j \neq 0\quad &\Rightarrow\quad X_j \notindependent Y\, |\, X_G.
\end{align}
For cross-entropy, the converse holds as well.
\end{result}
\begin{proof}
    For L2-optimal models, $\fh^c_\emptyset = \E[Y] = \fh^r_\emptyset$ and $\fh^c_G = \E[Y|X_G] = \fh^r_G$. For cross-entropy optimal models, $\fh^c_\emptyset = \P(Y) = \fh^r_\emptyset$ and $\fh^c_G = \P(Y|X_G) = \fh^r_G$. Thus, the interpretation is the same as for cSAGEvf (Result \ref{cSAGEvf_int}). \qed
\end{proof}

\section{Examples} \label{sec_example}

We can now answer the open questions of the motivational example from the introduction (Section \ref{intro}). 
To illustrate our recommendations (summarized in Table \ref{tab_summary_results}), we additionally apply the FI methods to a simplified setting where the DGP and the model's mechanism are known and intelligible, including features with different roles.
\begin{table}[t]
\caption{Summary of our results. The abbreviation ``CE" stands for cross-entropy loss and ``L2" for L2-loss, each with the respective optimal model. \label{tab_summary_results}}
    \begin{center}
         \begin{tabular}{p{2.5cm} l l } 
         \toprule
         \textbf{Outcome} & \textbf{Assumptions} & \textbf{Implication} \\
         \hline 
         $\text{PFI}_j \neq 0$ & $X_{j} \independent X_{-j}\, | \,Y$ & $\Rightarrow \quad X_j \notindependent Y$ \\
         $ \text{PFI}_j = 0$&  CE $\land\, (X_j \independent X_{-j}) \land (X_j \independent X_{-j}\, |\, Y)$ & $\Rightarrow \quad X_j \independent Y$ \\
         mSAGEvf$_j \neq 0$ & (L2 $\lor$ CE) $\land\, (X_j \independent X_{-j})$ & $\Rightarrow \quad X_j \notindependent Y$ \\
         mSAGEvf$_j$ $ = 0$ & CE $\land\, (X_j \independent X_{-j})$ &  $\Rightarrow \quad X_j \independent Y$ \\
         cSAGEvf$_j$ $\neq 0$ & L2 $\lor$ CE & $\Rightarrow \quad X_j \notindependent Y$ \\
         cSAGEvf$_j$ $ = 0$ & CE & $\Rightarrow \quad X_j \independent Y$ \\
         $\text{LOCI}_{j} \neq 0$ & L2 $\lor$ CE & $\Rightarrow \quad X_j \notindependent Y$ \\
         $\text{LOCI}_{j} = 0$ & CE & $\Rightarrow \quad X_j \independent Y$ \\
         \hline 
         $\text{CFI}_j \neq 0$ & - & $\Rightarrow \quad X_j \notindependent Y \, |\, X_{-j}$ \\
         $ \text{CFI}_j = 0$ & CE & $\Rightarrow \quad X_j \independent Y \, |\, X_{-j}$ \\
         scSAGEvf$^{-j}_j \neq 0$ & L2 $\lor$ CE & $\Rightarrow \quad X_j \notindependent Y \, |\, X_{-j}$ \\
         scSAGEvf$^{-j}_j = 0$ & CE & $\Rightarrow \quad X_j \independent Y \, |\, X_{-j}$ \\
         $\text{LOCO}_j \neq 0$ & L2 $\lor$ CE & $\Rightarrow \quad X_j \notindependent Y \, |\, X_{-j}$ \\
         $\text{LOCO}_{j} = 0$ & CE & $\Rightarrow \quad X_j \independent Y \, |\, X_{-j}$ \\
         \hline 
         $\text{RFI}_j^G \neq 0$ & $X_j \independent X_R \, |\, X_G, Y$ & $\Rightarrow \quad X_j \notindependent Y \, |\, X_{G}$ \\
         $\text{RFI}^G_j = 0$ & CE $\land\, (X_j \independent X_R \, |\, X_G, Y) \land (X_j \independent X_R \, |\, X_G)\quad$ & $\Rightarrow \quad X_j \independent Y \, |\, X_{G}$ \\
         scSAGEvf$^G_j \neq 0$ & L2 $\lor$ CE & $\Rightarrow \quad X_j \notindependent Y \, |\, X_{G}$ \\
         scSAGEvf$^G_j = 0$ & CE & $\Rightarrow \quad X_j \independent Y \, |\, X_{G}$ \\
         $\text{sWVIM}^{-G}_j \neq 0$ & L2 $\lor$ CE & $\Rightarrow \quad X_j \notindependent Y \, |\, X_{G}$ \\
         $\text{sWVIM}^{-G}_j = 0$ & CE & $\Rightarrow \quad X_j \independent Y \, |\, X_{G}$ \\
         \bottomrule
        \end{tabular} 
    \end{center}
\end{table} 
\subsubsection{Returning to our Motivating Example.}{Using Result \ref{result:non-zero-pfi}, we know that PFI can assign high FI values to features even if they are not associated with the target but with other features that are associated with the target. 
Conversely, LOCO only assigns non-zero values to features conditionally associated with the target (here: bike rentals per day, see Result \ref{result:loco-interpretation}). 
We can therefore conclude that at least the features \texttt{weathersit, season, temp, mnth, windspeed} and \texttt{weekday} are conditionally associated with the target, and the TOP 5 most important features, according to PFI, tend to share information with other features or may not be associated with \texttt{bike rentals per day} at all.}
\subsubsection{Illustrative Example with known Ground-truth.}{This example includes five features $X_1, \dots, X_5$ and a target $Y$ with the following dependence structure (visualized in Figure \ref{fig:ill_examp_FI}, left plot):
\begin{itemize}
    \item $X_1, X_3$ and $X_5$ are independent and standard normal: $X_j \sim N(0,1)$,
    \item $X_2$ is a noisy copy of $X_1$: $X_2 := X_1 + \epsilon_2, \epsilon_2 \sim N(0, 0.001)$,
    \item $X_4$ is a (more) noisy copy of $X_3$: $X_4 := X_3 + \epsilon_4, \epsilon_4 \sim N(0, 0.1)$,
    \item 
    $Y$ depends on $X_4$ and $X_5$ via linear effects and a bivariate interaction:\\
    $Y := X_4 + X_5 + X_4*X_5 + \epsilon_Y,\, \epsilon_Y \sim N(0, 0.1).$
\end{itemize}
Regarding \ref{question:unconditional}, features $X_3, X_4$ and $X_5$ are unconditionally associated with $Y$, while only  $X_5$ is conditionally associated with $Y$ given all other features \ref{question:conditional}.

We sample $n = 10,000$ observations from the DGP and use 70\% of the observations to train two models: 
A linear model (LM) with additional pair-wise interactions between all features (test-MSE $=0.0103$, test-$R^2=0.9966$), and a random forest (RF) using default hyperparameters (test-MSE $=0.0189$, test-$R^2=0.9937$).
We apply the FI methods on 30\% test data with L2 loss to both models using 50 repetitions for methods that marginalize or perturb features.
We present the results in Figure \ref{fig:ill_examp_FI}.\footnote{All FI methods and reproducible scripts for the experiments are available online via \url{https://github.com/slds-lmu/paper_2024_guide_fi.git}.
Most FI methods were computed with the Python package \texttt{fippy} (\url{https://github.com/gcskoenig/fippy.git}).}
The right plot shows each feature's FI value relative to the most important feature (which is scaled to 1). 
\begin{figure}[t]
\begin{subfigure}[t]{.4\textwidth}
\footnotesize
    \begin{tikzpicture}[node distance = 0.3cm, scale=0.83,every node/.style={scale=0.88, line width=0.25mm, black, fill=white}]

    \draw[white, fill=gray!15] (-3,-1.7) -- (3,-1.7) -- (3,0.55) -- (-3,0.55) -- cycle;

    \draw[white, fill=gray!15] (-3,-1.9) -- (3,-1.9) -- (3,-4) -- (-3,-4) -- cycle;
      \node [draw, circle] (yhat) at (0,0) {$\hat{Y}$};
      \node [draw, circle] (x33) [below = of yhat] {$X_3$};
      \node [draw, circle] (x22) [left = of x33 ] {$X_2$};
      \node [draw, circle] (x11) [left = of x22 ] {$X_1$};
      \node [draw, circle] (x44) [right = of x33] {$X_4$};
      \node [draw, circle] (x55) [right = of x44 ] {$X_5$};
      \node[scale=1] (ml) at (1.8,0.15) {model level};
     
     \draw[-] (yhat) -- (x44);
     \draw[-] (yhat) -- (x33);
     \draw[-] (yhat) -- (x22);
     \draw[-] (yhat) -- (x11);
     \draw[-] (yhat) -- (x55);

     \node [draw, circle] (x4) [below = of x44] {$X_4$};
      \node [draw, circle] (x3) [left = of x4 ] {$X_3$};
      \node [draw, circle] (x2) [left = of x3 ] {$X_2$};
      \node [draw, circle] (x1) [left = of x2 ] {$X_1$};
      \node [draw, circle] (x5) [right = of x4 ] {$X_5$};
      \node [draw, circle] (y) [below = of x3] {$Y$};
      \node[scale=1] (dl) at (1.8,-3.6) {data level};

      \draw[-] (x2) -- (x1);
        \draw[-] (x3) -- (x4);
		\draw[-] (x4) -- (y);
        \draw[-] (x5) -- (y);

        \draw[-, dotted] (x1) -- (x11);
        \draw[-, dotted] (x2) -- (x22);
        \draw[-, dotted] (x3) -- (x33);
        \draw[-, dotted] (x4) -- (x44);
        \draw[-, dotted] (x5) -- (x55);

        \draw[white, fill=white] (-3,-5.7) -- (3,-5.7) -- (3,-5.8) -- (-3,-5.8) -- cycle;
    
    \end{tikzpicture}
\end{subfigure}\hfill%
\begin{subfigure}[t]{.59\textwidth}
  \includegraphics[clip, trim=0.55cm 0cm 0.1cm 0cm, width=\linewidth]{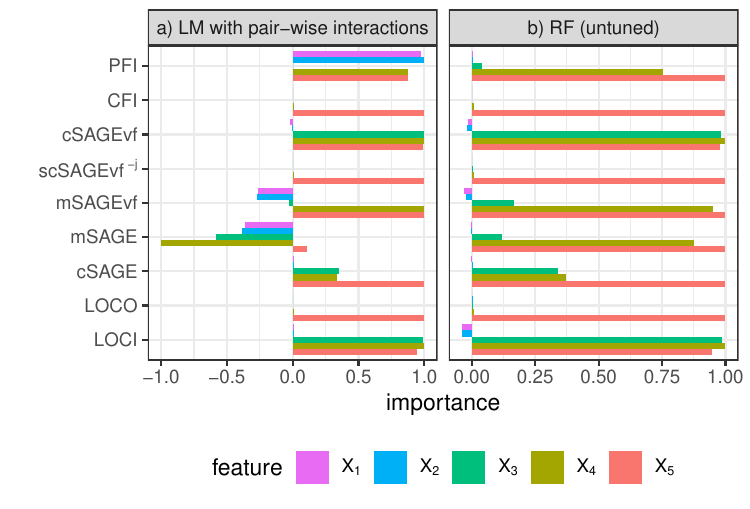}
\end{subfigure}
  \caption{Left: Graph illustrating the model and data level associations. Right: Results of FI methods for the LM in panel (a) and the RF in panel (b); importance values are relative to the most important feature.}  \label{fig:ill_examp_FI}
\end{figure}
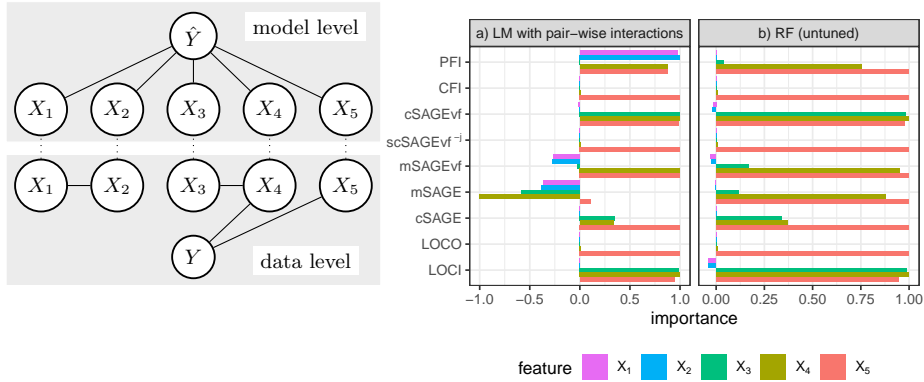
%

\paragraph{\ref{question:unconditional}:} \textit{LOCI} and \textit{cSAGEvf} correctly identify $X_3$, $X_4$ and $X_5$ as unconditionally associated. 
\textit{PFI} correctly identifies $X_4$ and $X_5$ to be relevant, but it misses $X_3$, presumably since the model predominantly relies on $X_4$. 
For the LM, \textit{PFI} additionally considers $X_1$ and $X_2$ to be relevant, although they are fully independent of $Y$; due to correlation in the feature sets, the trained model includes the term $0.36x_1-0.36x_2$, which cancels out in the unperturbed, original distribution, but causes performance drops when the dependence between $X_1$ and $X_2$ is broken via perturbation. 
For \textit{mSAGEvf}, similar observations can be made, with the difference that $X_1$ and $X_2$ receive negative importance. 
The reason is that for \textit{mSAGEvf}, the performance of the average prediction is compared to the prediction where all but one feature are marginalized out; 
we would expect that adding a feature improves the performance, but for $X_1$ and $X_2$, the performance worsens if adding the feature breaks the dependence between $X_1$ and $X_2$. 

\paragraph{\ref{question:conditional-general}:} \textit{CFI}, \textit{LOCO}, and \textit{scSAGEvf$^{-j}$} correctly identify $X_5$ as conditionally associated, as expected.
\textit{cSAGE} correctly identifies features that are dependent with $Y$ conditional on any set $S$, specifically, $X_3$, $X_4$ and $X_5$. The results of \textit{mSAGE} for the RF are similar to those for \textit{cSAGE}; on the LM, the results are quite inconclusive -- most features have a negative importance.

Overall, the example empirically illustrates the differences between the methods as theoretically shown in Sections \ref{subsec:rules:univariate} to \ref{subsec:rules:refitting}.

\section{Summary and Practical Considerations} \label{recommendations}

In Sections \ref{subsec:rules:univariate} to \ref{subsec:rules:refitting}, we presented three different classes of FI techniques: Techniques based on univariate perturbations, techniques based on marginalization, and techniques based on model refitting. 
In principle, each approach can be used to gain partial insights into questions \ref{question:unconditional} to \ref{question:relative}. 
However, the practicality of the methods depends on the specific application. 
As follows, we discuss some aspects that may be relevant to the practitioner.

For \ref{question:unconditional}, PFI, mSAGEvf, cSAGEvf, and LOCI are -- in theory -- suitable. 
However, PFI and mSAGEvf require assumptions about feature independence, which are typically unrealistic.
cSAGEvf require marginalizing out features using a multivariate conditional distribution $P(X_{-j}|X_j)$, which can be challenging since not only the dependencies between $X_j$ and $X_{-j}$ but also the ones between $X_{-j}$ have to be considered.
LOCI requires fitting a univariate model, which is computationally much less demanding than the cSAGEvf computation.

For \ref{question:conditional}, a comparatively more challenging task, CFI, scSAGEvf and LOCO are suitable, but it is unclear which of the methods is preferable in practice. 
While CFI and scSAGEvf require a model of the univariate conditional $P(X_j|X_{-j})$, LOCO requires fitting a model to predict $Y$ from $X_{-j}$.
For \ref{question:relative}, the practical requirements depend on the size of the conditioning set. 
The closer the conditioning set is to $-j$, the fewer features have to be marginalized out for scSAGEvf, and the fewer feature dependencies may lead to extrapolation for RFI. 
For sWVIM, larger relative feature sets imply more expensive model fits.

Importantly, all three questions \ref{question:unconditional} to \ref{question:relative} could also be assessed with direct or conditional independence tests, e.g., mutual information \cite{cover1999elements}, partial correlation tests \cite{baba2004partial}, kernel-based measures such as the Hilbert-Schmidt independence criterion \cite{Gretton2005,zhang2012kernel}, or the generalized covariance \cite{shah2020hardness}. 
This seems particularly appropriate for question \ref{question:unconditional}, where we simply model the association structure of a bivariate distribution. 
Methods like mSAGEvf can arguably be considered overly complex and computationally expensive for such a task.}

\section{Statistical Inference for FI Methods} \label{uncertainty}

So far, we have described how the presented FI methods should behave in theory or as point estimators.
However, the estimation of FI values is inherently subject to various sources of uncertainty introduced during the FI estimation procedure, model training, or model selection \cite{molnar2023relating,Williamson2023}. 
This section reviews available techniques to account for uncertainty in FI by applying methods of statistical inference, e.g., statistical tests and the estimation of confidence intervals (CIs). 

All FI methods in this paper measure the expected loss. 
To prevent biased or misleading estimates due to overfitting, it is crucial to calculate FI values on independent test data not seen during training, aligning with best practices in ML performance assessment \cite{simon2007,Lones2021}. 
Computing FI values on training data may lead to wrong conclusions. For example, Molnar et al. \cite{Molnar2022} demonstrated that even if features are random noise and not associated with the target, some features are incorrectly deemed important when FI values are computed using training data instead of test data. 
If no large dedicated test set is available, or the data set is not large in general to facilitate simple holdout splitting,   
resampling techniques such as cross-validation or bootstrap provide practical solutions \cite{simon2007}.

In the following, we will first provide an overview of method-specific approaches and then summarize further ideas about more general ones. 

\textit{PFI and CFI.}
Molnar et al. \cite{molnar2023relating} address the uncertainty of model-specific PFI and CFI values caused by estimating expected values using Monte Carlo integration on a fixed test data set and model. 
To address the variance of the learning algorithm, they introduce the learner-PFI, computed using resampling techniques such as bootstrapping or subsampling on a held-out test set within each resampling iteration. 
They also propose variance-corrected Wald-type CIs to compensate for the underestimation of variance caused by partially sharing training data between the models fitted in each resampling iteration.
For CFI, Watson and Wright \cite{Watson2021} address sampling uncertainty by comparing instance-wise loss values. 
They use Fisher's exact (permutation) tests and paired $t$-tests for hypothesis testing. 
The latter, based on the central limit theorem, is applied to all decomposable loss functions calculated by averaging instance-wise losses. 

\textit{SAGE.} The original paper of SAGE \cite{Covert2020} introduced an efficient algorithm to approximate SAGE values, since the exact calculation of SAGE values is computationally expensive. 
They show that, according to the central limit theorem, the approximation algorithm convergences to the correct values and that the variance reduces with the number of iterations at a linear rate. 
They briefly mention that the variance of the approximation can be estimated at a specific iteration and can be used to construct CIs (which corresponds to the same underlying idea of the Wald-type CI for the model-specific PFI mentioned earlier).

\textit{WVIM including LOCO.}
Lei et al. \cite{Lei2018} introduced statistical inference for LOCO by splitting the data into two parts: one for model fitting and one for estimating LOCO.
They further employed hypothesis testing and constructing CIs using sign tests or the Wilcoxon signed-rank test. 
The results' interpretation is limited to the importance of the FOI to an ML algorithm's estimated model on a fixed training data set.
Williamson et al. \cite{Williamson2023} construct Wald-type CI intervals for LOCO and WVIM, based on $k$-fold cross-validation and sample-splitting\footnote{This involves dividing the $k$-folds into two parts to serve distinct purposes, allowing for separate estimation and testing procedures.}. 
Compared to LOCO, it provides a more general interpretation of the results as it considers the FI of an ML algorithm trained on samples of a particular size, i.e., due to cross-validation, the results are not tied to a single training data set.
The approach is related to \cite{molnar2023relating} but removes features via refitting instead of sampling and does not consider any variance correction.
The authors note that, while sample-splitting helps to address issues related to zero-importance features having an incorrect type I error or coverage of their CIs, it may not fully leverage all available information in the data set to train a model.
\paragraph{PIMP.} The PIMP heuristic \cite{Altmann2010} is based on model refits and was 
initially developed to address bias in FI measures such as PFI within random forests.
However, PIMP is a general procedure and has broader applicability across various FI methods \cite{linardatos2020explainable,Molnar2022}. 
PIMP involves repeatedly permuting the target to disrupt its associations with features while preserving feature dependencies, training a model on the data with the permuted target, and computing PFI values.
This leads to a collection of PFI values (called null importances) under the assumption of no association between the FOI and the target.
The PFI value of the model trained on the original data is then compared with the distribution of null importances to identify significant features.
%
\paragraph{Methods Based on the Rashomon Set.}
The Rashomon set refers to a collection of models that perform equally well but may differ in how they construct the prediction function and the features they rely on.
Fisher et al. \cite{Fisher2019} consider the Rashomon set of a specific model class (e.g., decision trees) defined based on a performance threshold and propose a method to measure the FI within this set.
For each model in the Rashomon set, the FI of a FOI is computed, and its range across all models is reported. 
Other works include the Variable Importance Cloud (VIC) \cite{Dong2019}, providing a visual representation of FI values over different model types; the Rashomon Importance Distribution (RID) \cite{donnelly2024rashomon}, providing the FI distribution across the set and CIs to characterize uncertainty around FI point estimates; and ShapleyVIC \cite{Ning2022}, extending VIC to SAGE values and using a variance estimator for constructing CIs. 
The main idea is to address uncertainty in model selection by analyzing a Rashomon set, hoping that some of these models reflect the underlying DGP and assign similar FI values to features.
\paragraph{Multiple Comparisons.} Testing multiple FI values simultaneously poses a challenge known as multiple comparisons.
The risk of falsely rejecting true null hypotheses increases with the number of comparisons. 
Practitioners can mitigate it, e.g., by controlling the family-wise error rate or the false discovery rate \cite{Romano2016,Molnar2022}.

\section{Open Challenges and Further Research} \label{conclusion}

\paragraph{Feature Interactions.}
FI computations are usually complicated by the presence of strong and higher-order interactions~\cite{Molnar2022}. 
Such interactions typically have to be manually specified in (semi-)parametric statistical models. However, complex non-parametric ML models, to which we usually apply our model-agnostic IML techniques, automatically include higher-order interaction effects.
While recent advances have been made in visualizing the effect of feature interactions and quantifying their contribution regarding the prediction function~\cite{apley2020visualizing,Greenwell2018,herbinger2023decomposing}, we feel that this topic is somewhat underexplored in the context of loss-based FI methods, i.e., how much an interaction contributes to the predictive performance.
A notable exception is SAGE, which, however, does not explicitly quantify the contribution of interactions towards the predictive performance but rather distributes interaction importance evenly among all interacting features.
In future work, this could be extended by combining ideas from functional decomposition~\cite{apley2020visualizing,herbinger2023decomposing}, FI based on those~\cite{hiabu2023unifying} and loss-based methods as in SAGE. 
\paragraph{Model Selection and AutoML.} 
As a subtle but important point: it seems somewhat unclear to which model class or learning algorithms the covered techniques can or should be applied to, if DGP inference is the goal. From a mechanistic perspective, these model-agnostic FI approaches can be applied to basically any model class, which seems to be the case in current applications.
Considering what Williamson et al. \cite{Williamson2023} noted in and, following our results, many statements in the Sections \ref{subsec:rules:univariate} to \ref{subsec:rules:refitting} only hold under a ``loss-optimal model''.
First of all, in practice, the construction of a loss-optimal model with certainty is virtually impossible. 
Does this imply we should try to squeeze out as much predictive performance as possible, regardless of the incurred extra model complexity? 
Williamson et al. \cite{Williamson2023} use the ``super learner'' in their definition and implementation of WVIM \cite{vimp}.
Modern AutoML systems like AutoGluon~\cite{erickson2020autogluon} are based on the same principle. 
While we perfectly understand that choice, and find the combination of AutoML and IML techniques very exciting, we are unsure about the trade-off costs. 
Certainly, this is a computationally expensive technique. 
But we rather also worry about the underlying implications for FI methods (or more generally IML techniques), when models of basically the highest order of complexity are now used, which usually contain nearly unconstrained higher-order interactions. 
We think that this issue needs to be more analyzed.
\paragraph{Rashomon Sets and Model Diagnosis.}
Expanding on the previous issue: 
In classical statistical modeling, models are usually not exclusively validated by checking predictive performance metrics only. The Rashomon effect tells us that in quite a few scenarios, very similarly performing models exist, which give rise to different response surfaces and different IML interpretations. 
This hints at the effect that ML researchers and data scientists might likely have to expand their model validation toolbox, in order to have better options to exclude misspecified models.
\paragraph{Empirical Performance Comparisons.} 
We have tried to compile a succinct list of results to describe what can be derived from various FI methods regarding the DGP.
However, we would also like to note that such theoretical analysis often considerably simplifies the complexity of real-world scenarios to which we apply these techniques.
For that reason, it is usually a good idea to complement such mathematical analysis with informative, detailed, and carefully constructed empirical benchmarks. 
Unfortunately, not a lot of work on empirical benchmarks exists in this area. 
Admittedly, this is not easy in FI, as ground truths are often only available in simulations, which, in turn, lack the complexity found in real-world data sets.
Moreover, even in simulations, concrete ``importance ground truth numbers'' might be debatable. 
So far, there are no extensive benchmarks in the literature on FI methods. 
Many compare local importance methods \cite{Agarwal2022,Han2022}, but few global methods: 
E.g., Blesch et al. \cite{Blesch2023} and Covert et al. \cite{Covert2020} compare FI methods for different data sets, metrics, and ML models. 
However, the comparisons are not applied with regard to different association types, as the methods are not differentiated in this respect as in our paper.
\paragraph{Causality.} {Beyond association, scientific practitioners are often interested in causation (see, e.g., \cite{yazdani2015causal,varian2016causal,glass2013causal,gangl2010causal,rothman2005causation}). 
In our example from Section \ref{sec:scope_DGP_properties}, the doctor may not only want to predict the disease but may also want to treat it. 
Knowing which features are associated with the disease is insufficient for that purpose -- association remains on rung 1 of the so-called ladder of causation \cite{pearl2018book}:
Although the symptoms are associated with the disease, treating them does not affect the disease.
To gain insight into the effects of interventions (rung 2), experiments and/or causal knowledge and specialized tools are required \cite{pearl2009causality,imbens2015causal,peters2017elements,hernan2023whatif}.

\begin{credits}
\subsubsection{\ackname}
MNW was supported by the German Research Foundation (DFG), Grant Numbers: 437611051, 459360854.
GK was supported by the German Research Foundation through the Cluster of Excellence “Machine Learning - New Perspectives for Science" (EXC 2064/1 number 390727645).

\subsubsection{\discintname}
The authors have no competing interests to declare that are relevant to the content of this article. 
\end{credits}

\section*{Appendix}
\appendix

\section{Additional proofs} \label{appendix_additional_proofs}

\subsection{Proof of Result \ref{result:zero-rfi-ce}}
\label{proof:result:zero-rfi-ce}
\begin{proof}
    We show that $X_j \notindependent Y\, |\, X_G \Rightarrow RFI_j^G \neq 0$: 
    For cross-entropy loss,
    \begin{align*}
        RFI_j &= (E_X[D_{KL}(p(y|x) || f(y|x_{-j}, \tilde{x}_j))] - H(Y|X))\\
        &\quad - (E_X[D_{KL}(p(y|x) || f(y|x))] - H(Y|X))\\
        &\overset{f=p}{=} E_X[D_{KL}(p(y|x) || f(y|x_{-j}, \tilde{x}_j)]
    \end{align*}
    It remains to show that KL-divergence for $f(y, x_{-j}, \tilde{x}_j)$ is non-zero:
    \begin{align*}
        p(x_j, y, x_G, x_R) &= p(x_j|y, x_G, x_R)\\
        &= p(x_j|y, x_G) p(y, x_G, x_R) & & (X_j \independent X_R | X_G, Y)\\
        &\neq p(x_j|x_G) p(y, x_G, x_R) & & (X_j \notindependent Y\, |\, X_G) \\
        &= p(\tilde{x_j}|x_G) p(y, x_G, x_R) && (\text{def. of }\tilde{X}_j)\\
        &= p(\tilde{x_j}, y, x_G, x_R)
    \end{align*}
    Since $X_j \independent X_R | X_G$ it holds that $p(\tilde{x}_j, x_R, x_G) = p(x_j, x_R, x_G)$ and, thus, $p(y|\tilde{x}_j, x_R, x_G) \neq p(y|x_j, x_R, x_G)$.
    With model optimality, $p(y|x) \neq f(y|\tilde{x}_j, x_{-j})$.
    Since KL divergence $> 0$ for $p \neq f$ it holds that $RFI_j > 0$.
    \qed
\end{proof}

\subsection{Proof of Result \ref{mSAGEvf_int}: mSAGEvf interpretation} \label{proof_eq_mSAGEvf}
\begin{proof}
The implication is shown by proving the counterposition: 
$$X_j \independent (Y, X_{-j})\quad \Rightarrow \quad v^m(\{j\}) = 0.$$
Since $X_j \independent (Y, X_{-j}) \Rightarrow f^{\ast, m}_j(x_j) = f^{\ast, c}_j(x_j)$ it holds that $v^m(\{j\}) = v^c(\{j\})$.
$X_j \independent (Y, X_{-j}) \Rightarrow X_j \independent Y$ and thus $v^m(\{j\}) = v^c(\{j\}) = 0$ (Result \ref{cSAGEvf_int}). \qed
\end{proof}

\subsection{Proof of Result \ref{cSAGE_interpretation}: cSAGE interpretation} \label{proof_eq_cSAGE}
\begin{proof}
The equation is shown by proving the contraposition
\begin{align*}
    \forall \SsubP\backslash\{j\}: X_j \independent Y \, |\, X_S \quad \Rightarrow \quad \phi^c_j(v) = 0. 
\end{align*}
From Result \ref{cSAGEvf_int} we know that $X_j \independent Y\, |\, X_G \Rightarrow v^c(G \cup \{j\}) - v^c(G) = 0$ for L2 and cross-entropy optimal predictors. If $\forall \SsubP\backslash\{j\}: X_j \independent Y \, |\, X_S$, all summands of the SAGE value are zero, and thus $\phi_j^c = 0$.

\textit{Converse for cross-entropy loss:}
%
We prove the converse by counterposition
\begin{align*}
    \phi^c_j(v) = 0 \quad \Rightarrow \quad \forall \SsubP\backslash\{j\}: X_j \independent Y \, |\, X_S.
\end{align*}
If $L$ is the cross-entropy loss and $\fbayes$ the Bayes model, using \cite[Appendix C.1]{Covert2020}
$$\phi^c_j(v) = \frac{1}{p} \sum_{\SsubPnoj} \binom{p-1}{|S|}^{-1} I(Y; X_{j}\,|\, X_S) = 0,$$
where the mutual information $I$ and the coefficients are always non-negative. Thus, we add non-negative terms so the sum can only be zero if
$\forall \SsubP\backslash\{j\}: I(Y; X_{j}\,|\, X_S) = 0$ and, thus, $\forall \SsubP\backslash\{j\}: X_j \independent Y \,|\, X_S.$\qed
\end{proof}

\bibliographystyle{splncs04}
\bibliography{lit/feature_importance}

\end{document}

%% file: z_basic-math.tex
\ifdefined\N                                                                
\renewcommand{\N}{\mathds{N}} 
\else \newcommand{\N}{\mathds{N}} \fi 
\newcommand{\R}{\mathds{R}} 
\ifdefined\C 
  \renewcommand{\C}{\mathds{C}} 
\else \newcommand{\C}{\mathds{C}} \fi

\renewcommand{\P}{\mathds{P}} 
\newcommand{\E}{\mathds{E}} 

%% file: z_basic-ml.tex
\newcommand{\Xspace}{\mathcal{X}} 
\newcommand{\Yspace}{\mathcal{Y}} 
\newcommand{\xv}{\mathbf{x}} 
\newcommand{\yv}{\mathbf{y}} 
\newcommand{\D}{\mathcal{D}} 
\newcommand{\xyi}[1][i]{\left(\xv^{(#1)}, y^{(#1)}\right)} 
\newcommand{\Dset}{\left( \xyi[1], \ldots, \xyi[n]\right)} 
\newcommand{\yvec}{\left(y^{(1)}, \hdots, y^{(n)}\right)^\top} 
\renewcommand{\xi}[1][i]{\xv^{(#1)}} 

\newcommand{\ftrue}{f_{\text{true}}}  
\newcommand{\Hspace}{\mathcal{H}} 
\newcommand{\fbayes}{f^{\ast}} 
\newcommand{\fh}{\hat{f}} 
\newcommand{\risk}{\mathcal{R}} 



%% file: z_ml-interpretable.tex
\newcommand{\pert}[3]{\ifthenelse{\equal{#2}{}}{\tilde{#1}}{\ifthenelse{\equal{#3}{}}{\tilde{#1}^{#2}}{\tilde{#1}^{#2|#3}}}}	

\newcommand{\fhnj}{\fh_{-j}} 

\newcommand{\Scupj}{S \cup \{j\}} 
\newcommand{\SsubP}{S \subseteq P} 
\newcommand{\SsubPnoj}{\SsubP \setminus \{j\}} 
